\documentclass[10pt,twocolumn,letterpaper]{article}

\usepackage[pagenumbers]{cvpr} 

\usepackage{graphicx}
\usepackage{amsmath}
\usepackage{amssymb}
\usepackage{booktabs}
\usepackage{units} 
\usepackage{times}
\usepackage{tabularx}
\usepackage{nicefrac}
\usepackage{enumitem}

\usepackage{caption}
\usepackage{xcolor}
\usepackage{mathtools}
\usepackage{xfrac}
\usepackage{multirow}
\usepackage{wasysym}
\usepackage{algorithm}
\usepackage{algpseudocode}

\usepackage[accsupp]{axessibility}  

\makeatletter

\@namedef{ver@everyshi.sty}{}
\newcommand\notsotiny{\@setfontsize\notsotiny{6.31415}{7.1828}} 

\makeatother

\usepackage{pgfplots}
\usepackage{tikz} 
\usetikzlibrary{patterns}
\pgfplotsset{compat=newest}

\usepackage[pagebackref,breaklinks,colorlinks]{hyperref}

\usepackage[capitalize]{cleveref}
\crefname{section}{Sec.}{Secs.}
\Crefname{section}{Section}{Sections}
\Crefname{table}{Table}{Tables}
\crefname{table}{Tab.}{Tabs.}

\newcommand{\smaller}{\fontsize{8.25}{8.25}\selectfont}
\newcommand{\probP}{\kern0.15em \text{I\kern-0.15em P}}
\newcommand{\FOG}{Fog~\cite{HahnerICCV21}}
\newcommand{\LISA}{LISA~\cite{LISA}}
\newcommand{\DROR}{DROR~\cite{DROR}}
\newcommand{\clear}{\textit{None}}
\newcommand{\wet}{\textvisiblespace \kern0.25em (ours)}

\newcommand{\both}{\textvisiblespace \kern-0.45em * (ours)}
\newcommand{\blank}[1]{\hspace*{#1}}

\DeclareMathOperator*{\argmax}{arg\,max}

\definecolor{water_color}{RGB}{69,156,238}
\definecolor{eth_orange}{RGB}{255,126,40}
\definecolor{eth_green}{RGB}{153,204,51}
\definecolor{eth_blue}{RGB}{169,204,242}
\definecolor{eth_red}{RGB}{230,140,132}
\definecolor{eth_gray}{RGB}{126,126,126}

\newcommand{\PAR}[1]{\vspace{-0.2eM}\vskip4pt \noindent{\bf #1}}

\def\cvprPaperID{1542} 
\def\confName{CVPR}
\def\confYear{2022}

\begin{document}

\title{LiDAR Snowfall Simulation for Robust 3D Object Detection}

\author{Martin Hahner $^1$\quad Christos Sakaridis $^1$\quad Mario Bijelic $^2$\\ 
Felix Heide $^{2}$\quad Fisher Yu $^{1}$\quad Dengxin Dai $^{1, 3}$\quad Luc Van Gool $^{1, 4}$\\[2mm] 
$^1$ ETH Zürich\quad $^2$ Princeton University \quad $^3$ MPI for Informatics \quad $^4$ KU Leuven
}
\maketitle

\begin{abstract}
\vspace{-0.7eM}
3D object detection is a central task for applications such as autonomous driving, in which the system needs to localize and classify surrounding traffic agents, even in the presence of adverse weather. In this paper, we address the problem of LiDAR-based 3D object detection under snowfall. Due to the difficulty of collecting and annotating training data in this setting, we propose a physically based method to simulate the effect of snowfall on real clear-weather LiDAR point clouds. Our method samples snow particles in 2D space for each LiDAR line and uses the induced geometry to modify the measurement for each LiDAR beam accordingly. Moreover, as snowfall often causes wetness on the ground, we also simulate ground wetness on LiDAR point clouds. We use our simulation to generate partially synthetic snowy LiDAR data and leverage these data for training 3D object detection models that are robust to snowfall. We conduct an extensive evaluation using several state-of-the-art 3D object detection methods and show that our simulation consistently yields significant performance gains on the real snowy STF dataset compared to clear-weather baselines and competing simulation approaches, while not sacrificing performance in clear weather. Our code is available at \href{https://github.com/SysCV/LiDAR_snow_sim}{github.com/SysCV/LiDAR\_snow\_sim}.
\end{abstract} 

\vspace{-0.7eM}\section{Introduction}
\label{sec:intro}
\begin{figure}
    \centering
    \vspace{-0.5eM}
    \includegraphics[width=\columnwidth]{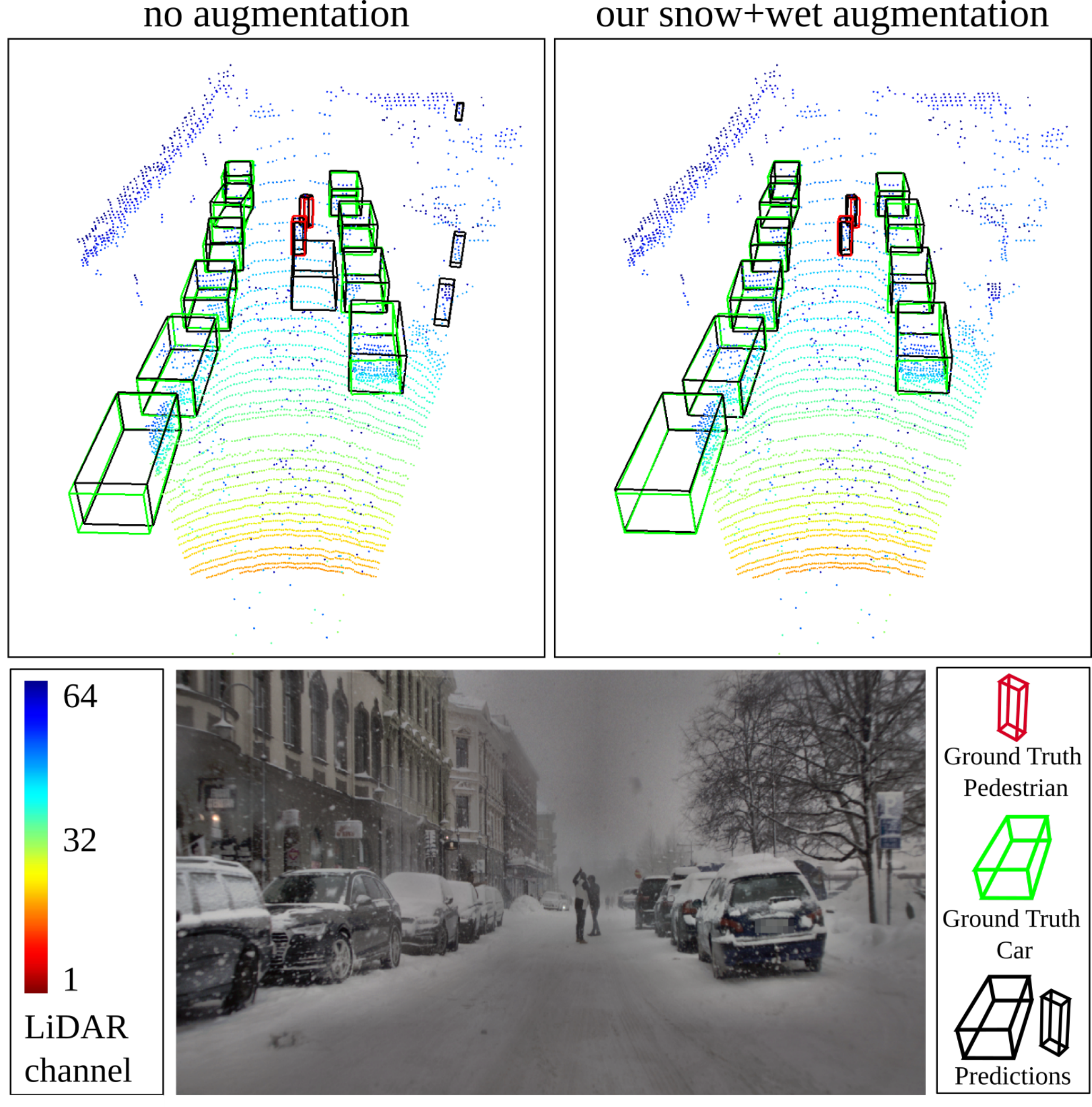}\vspace{-0.7eM}
    \caption{3D object detection results in heavy snowfall with prior training on the proposed data augmentation scheme (top right) in comparison to no augmentation (top left). The bottom row shows the RGB image as reference.
    }
    \label{fig:Teaser}\vspace{-1.6eM}
\end{figure}

A light detection and ranging (LiDAR) sensor is an active range sensor useful for several applications~\cite{PV-RCNN,tracking,localization,LOAM}. Its high-quality 3D output renders LiDAR the modality of choice for several tasks that require 3D reasoning, such as 3D object detection~\cite{PP,CP}. As LiDAR sensors are becoming increasingly cheaper~\cite{mems:mirrors:lidar:review}, their integration into autonomous cars becomes increasingly feasible as well. 

Nonetheless, previous sensor tests have revealed that such active pulsed systems are vulnerable in scattering media, leading to decreases of perception distances in various weather conditions such as rain~\cite{Goodin,LIBRE,Wallace}, fog~\cite{Heinzler,LIBRE,BenchmarkLidar,Jokela,Wallace}, and snow~\cite{Jokela, ArcticConditions,Michaud2015TowardsCT}, as shown in Fig.~\ref{fig:Teaser}. 
In these conditions, the optical medium contains particles of water or snow which interact with the laser beam and absorb, reflect or refract its photons. This results in two effects: (i) attenuation of the received power that corresponds to the target at the line of sight, and (ii) backscattering from particles leading to spurious maxima in the received power and thus to spurious returns at ranges different from the true range of the target. Consequently, there is a severe degradation of measurement quality due to intense noise, a large domain shift relative to point clouds captured in clear weather, and hence a detrimental effect on performance of high-level tasks such as 3D object detection~\cite{HahnerICCV21,STF}. Yet, achieving robust perception in adverse weather is a desirable goal as fatality rates for human drivers are notably higher in adverse weather, as reported by the US Department of Transportation~\cite{StaticsDeath} and the European Commission~\cite{EuropeanComission2018a}. 

Since adverse-weather data are hard to collect~\cite{STF}, previous works have investigated simulation methods to close the domain gap for camera data in fog~\cite{SDV18} and rain~\cite{Tremblay}. More recently, simulation methods for LiDAR sensors in fog~\cite{STF,HahnerICCV21} and rain~\cite{LISA,Goodin} have also been proposed. Motivated by this line of work, we introduce a physically based method to simulate snowfall on real clear-weather LiDAR point clouds. In particular, we use the linear system introduced in~\cite{Rasshofer_2011} to model the transmission of LiDAR pulses and the associated received power at the sensor. We simulate snowfall by explicitly sampling snow particles and modeling them as opaque spheres, the size of which is controlled by the snowfall rate~\cite{Canada1999,Gunn1958}. In our sampling, we obey the exclusion principle that no two particles intersect with each other. Given the sample of snow particles, we compute for each LiDAR beam the set of particles that intersect with it and derive the angle of the beam cross-section that is reflected by each particle, taking potential occlusions into account. This derivation directly delivers the modified impulse response of the linear system in the presence of snowfall, which allows the analytical calculation of the received power at the sensor.

Another condition associated with snowfall is wetness on the ground. This emerging thin water layer increases the specular component of reflection by the ground surface~\cite{WetRoadTest1}.
To model the ground reflection, we introduce an optical model using the Fresnel equations and the reflection on thin surfaces, which provides adapted reflectance values for wet surfaces.

The generated partially synthetic point clouds with our snowfall and wet ground simulation are used as training data for optimizing state-of-the-art 3D object detection methods, so that the learned models are more robust under snowfall. The hope is that our physically based simulation is realistic enough to relieve us from the need for real snowy training samples. We benchmark the models trained in this regime on the challenging real snowy subset of the STF dataset~\cite{STF} and find that the models trained on our simulated snow consistently achieve significant performance gains over baseline models trained only on clear weather and competing simulation methods.

\section{Related Work}
\label{sec:related}

\PAR{Adverse weather} research can be subdivided in meteorological publications providing fundamental knowledge for computer vision approaches \cite{Canada1999,Gunn1958,Sekhon1970,Rasshofer_2011}, phenomenological reasoning of introduced disturbance patterns of different weather conditions \cite{ArcticConditions, LIBRE, BenchmarkLidar} and the application of computer vision algorithms to such challenging conditions \cite{CADC,STF,ACDC,Tremblay}. In \cite{Rasshofer_2011} a general theoretical framework predicting the influence of various weather types on LiDAR has been studied, including rain, fog and snow.  
The authors study these weather effects following the statistical distribution of the scattering particles, which are introduced e.g.\ for snow in~\cite{Gunn1958} and \cite{Sekhon1970}. Further implications on visibility in snowfall are presented in~\cite{Canada1999}. 

Those resulting disturbance patterns and their strength are phenomenologically investigated for rain in \cite{Goodin,LIBRE,Wallace}, fog in \cite{Heinzler,LIBRE,BenchmarkLidar,Jokela,Wallace}, snow in\cite{Jokela, ArcticConditions,Michaud2015TowardsCT} and wet surfaces in \cite{Wojtanowski,WetAsphaltMeasurements,WetRoadTest1}. 
Algorithmically, authors have tried to tackle those conditions by robust fusion algorithms \cite{STF}, developed simulation techniques as data augmentation for camera data in \cite{Tremblay,SDV18} and LiDAR data in \cite{HahnerICCV21,LISA,STF,Goodin}. Authors also investigated enhancement technologies to remove adverse weather effects in \cite{ZeroScatter,DROR,PointCloudDeNoising2020} or applied domain adaptation methods adjusting clear-weather algorithms to adverse weather in \cite{ACDC}. 
Underlying deep learning data sets containing adverse weather samples were introduced in \cite{ACDC,CADC,STF,nuscenes2019,WaymoOD}. Yet classical data loop approaches are difficult to apply as adverse weather samples are rare and well underrepresented \cite{STF}.

\PAR{Simulation of adverse weather} allows to mitigate the rarity of adverse weather effects and difficulties in data collection campaigns as for example shown in \cite{SDV18,Tremblay,SynRealDataFogECCV18} through data synthesis. Additionally, it enables to generate reproducible conditions with clear ground truth necessary to learn image enhancement techniques in \cite{ZeroScatter,PointCloudDeNoising2020} or to investigate adverse weather noise-dependent performance decrease reproducible for different weather effects in \cite{Tremblay}. 
LiDAR simulation methods were explicitly studied in \cite{STF,HahnerICCV21,LISA,Goodin}. Developments started in \cite{STF} with a data driven approach in fog which was extended by \cite{HahnerICCV21}, introducing a physically based model and achieving higher 3D object detection performances. A simulation method for rain is introduced in \cite{Goodin} and a general approach for snow, rain and fog in \cite{LISA}. 
Contrary to~\cite{LISA}, our snowfall simulation involves a continuous formulation in the power signal domain, which allows us to superimpose reflections by different particles and to reason about occlusions between particles and the target, thereby adhering better to the physics of the laser transmission. Additionally, we take into account the effects of wet roads, allowing us to estimate laser hardware parameters as noise floor and sent intensity from the dry road intensities in a data-driven way.

\PAR{3D object detection} has seen tremendous progress in recent years. Several methods have been presented for RGB camera \cite{Simonelli_2019_ICCV,PseudoLidar,you2020pseudo}, LiDAR \cite{VoxelNet,PointNet,PP,PRCNN,PV-RCNN,shi2021pvrcnn,PartA2}, gated imagers \cite{Gated3D} or the fusion of multiple modalities in \cite{PointPainting,MV3D,AVOD,FrustumPointNet}. 
Across many dataset leaderboards, however, the top performance positions are typically all sorted out among LiDAR based methods \cite{WaymoOD,KITTI,nuscenes2019}.
In our work, we utilize the methods PV-RCCN~\cite{PV-RCNN}, VoxelRCNN-Car~\cite{VRCNN}, CenterPoint~\cite{CP}, Part-A²~\cite{PartA2}, PointRCNN~\cite{PRCNN}, SECOND~\cite{SECOND}, and PointPillars~\cite{PP}. The methods differ in point cloud representations, the used feature extraction backbones and the number of detection stages. As input modalities, point clouds are treated \eg in voxel space \cite{VoxelNet,SECOND,PV-RCNN,shi2021pvrcnn,PartA2}, inferring the raw point clouds \cite{PointNet,PRCNN} or using abstract representations such as pillars in \cite{PP}. 

The number of detection stages is most often classified into single \cite{PP} and two-stage approaches \cite{PRCNN,CP,PV-RCNN,VoxelNet,SECOND,shi2021pvrcnn}, where single-stage approaches directly discretize the input space and predict objects for each individual cell following \cite{SSD}. Two-stage approaches first predict proposals and refine them in a subsequent pooled feature space following the general idea of~\cite{FasterRCNN}. 

\section{Snowfall Simulation on LiDAR Point Clouds}
\label{sec:method}

\begin{figure}[!t]
     \centering
     \includegraphics[width=.8\linewidth]{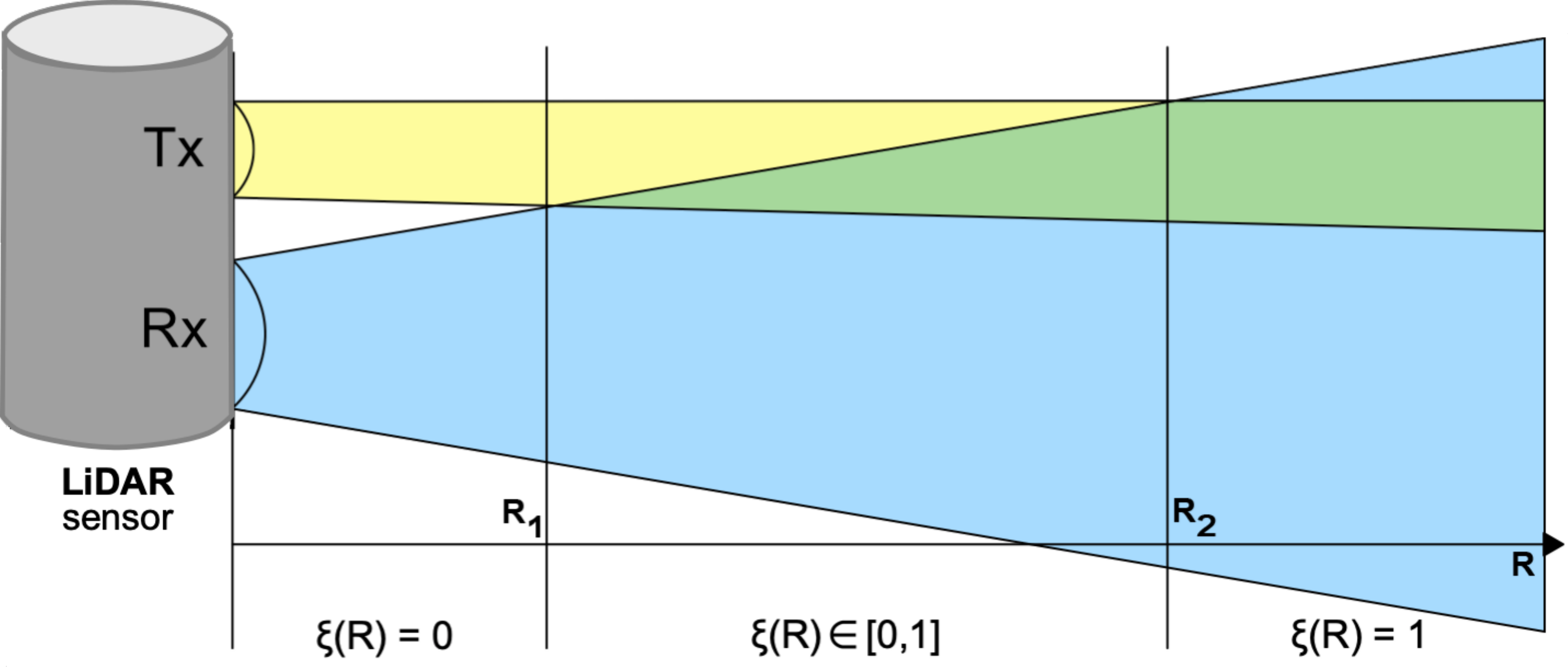}
     \vspace{-2mm}
     \caption{Sketch of a LiDAR sensor where the transmitter \textsf{Tx} and the receiver \textsf{Rx} do not have coaxial optics, but have parallel axes (called a bistatic beam configuration).
     }\vspace{-1eM}
     \label{fig:xi}
\end{figure}

\begin{figure*}
    \centering
    \includegraphics[width=\linewidth]{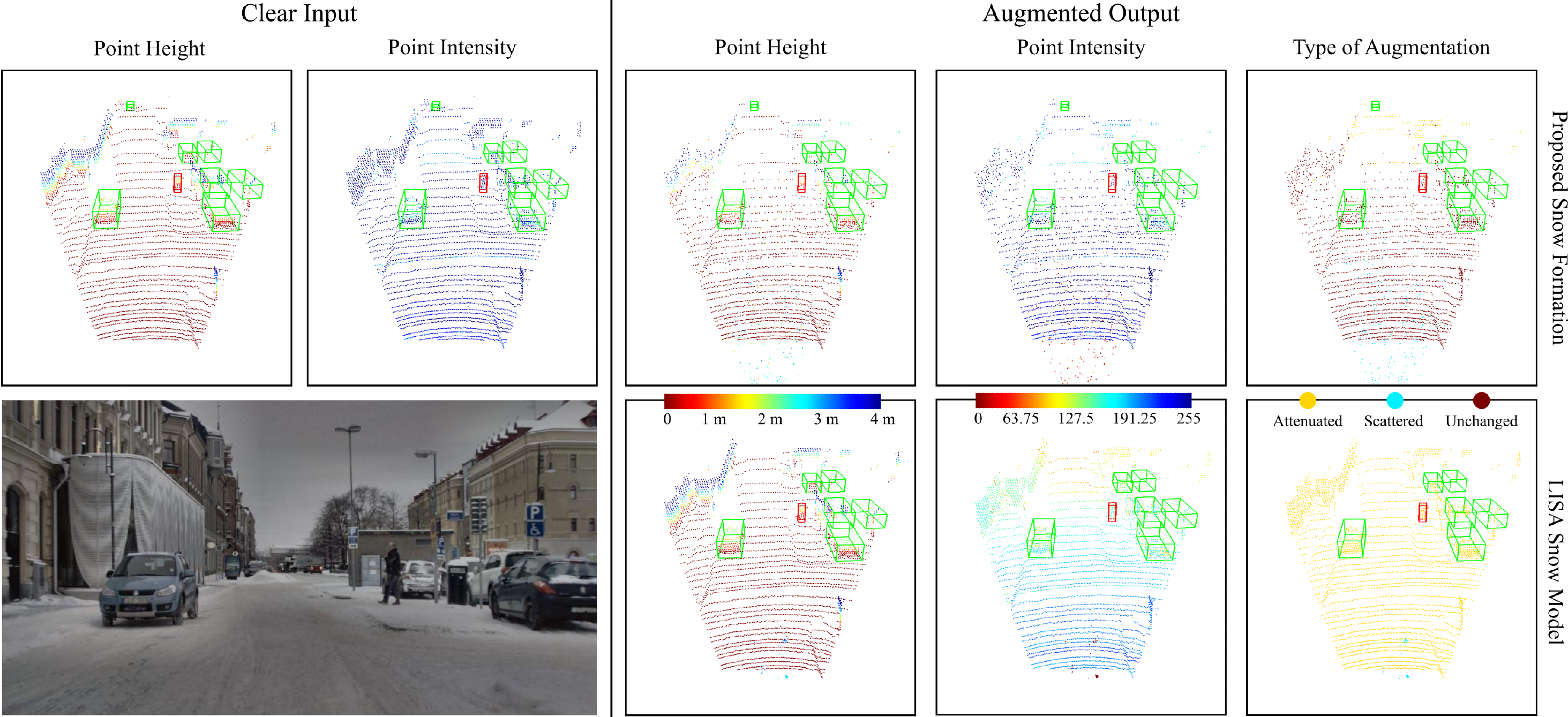} 
    \vspace{-6mm}
    \caption{Simulated snowfall corresponding to a snowfall rate of $r_s=\unit[2.5]{mm/h}$. The left block shows the clear undisturbed input. The right block shows our snowfall simulation (top) and the snowfall simulation in LISA~\cite{LISA} (bottom). Note that we simulate the scattering realistically and only attenuate points which are affected by individual snowflakes instead of attenuating all points based on their distance.}
    \label{fig:QualitativeResults}\vspace{-1.5eM}
\end{figure*}

\begin{figure}[!t]
    \centering
    \includegraphics[width=\linewidth]{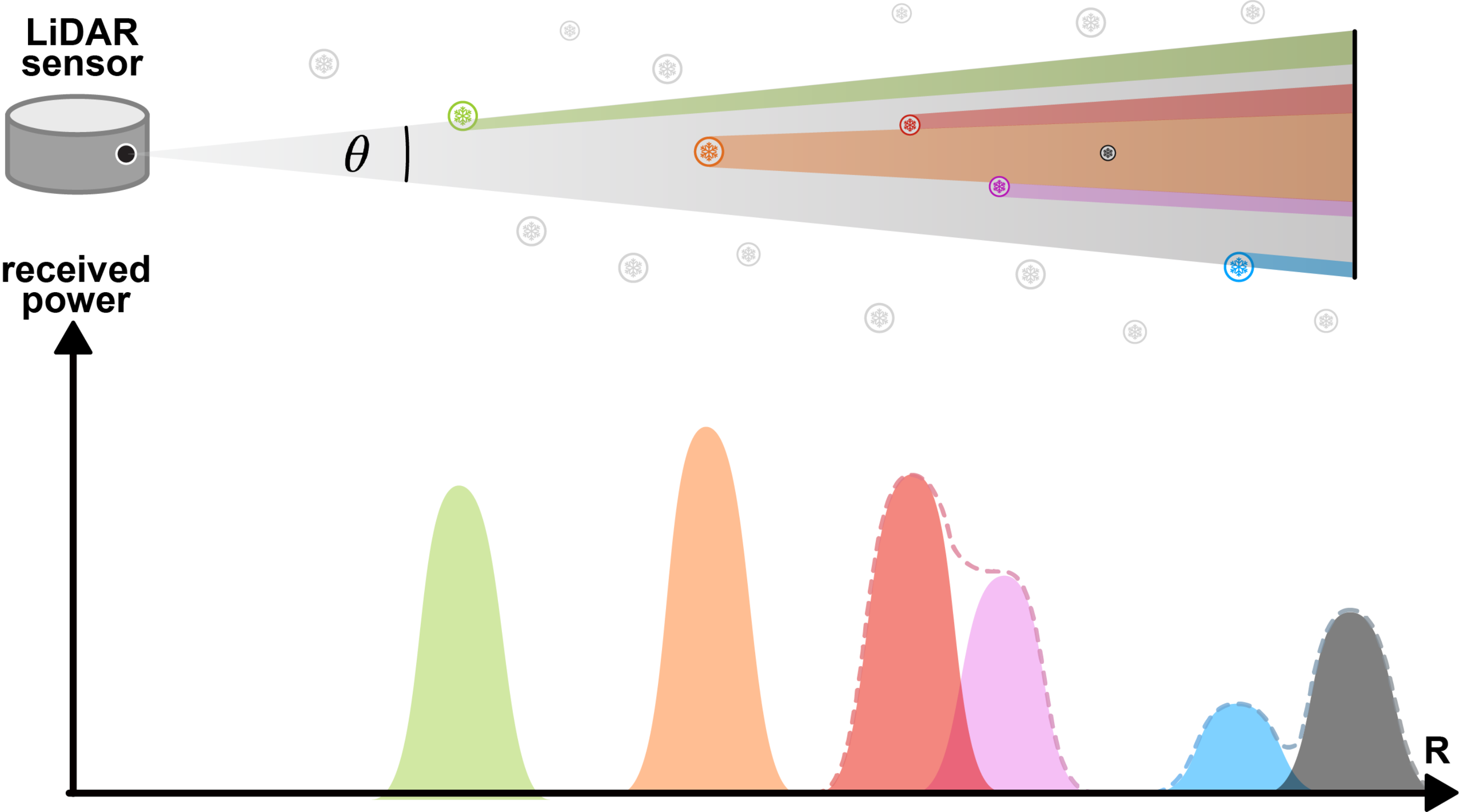}
    \vspace{-1.5eM}
    \caption{Snow particles interfering a single LiDAR beam (top).
    Schematic plot of corresponding received power echoes (bottom). Note how the received power of individual targets can overlap with each other ($c \tau_H \approx \unit[3]{m}$ with $\tau_H= \unit[10]{ns}$).}\vspace{-2eM}
    \label{fig:theory}
\end{figure}
 
\PAR{Pulse propagation in free space} can be modeled with geometrical optics for cost-effective LiDAR systems. Such systems apply an array of synchronized near-field infrared pulse emitters \textsf{Tx} and avalanche photodiodes (APDs) as receivers \textsf{Rx} depicted in Fig.~\ref{fig:xi} and described in \cite{EyeSafeLidar}. 
The sent-out laser pulse $P_0$ is reflected by a solid scene object, often referred to as target, with reflectivity $\rho_0$, and it is captured by the receiver, providing the time delay $\tau$ of the captured echo and its corresponding power $P_R$. The object distance $R$ is calculated by applying $R=c\tau$, where $c$ is the speed of light. The 3D position $[x,y,z]$ of the object is obtained by using the direction in which the pulse was emitted. For extended objects, geometric optics~\cite{GeometricalIntensityModel} can be applied to model the received power $P_R$ following
\begin{equation}\label{eq:rigid:sensormodel}
    \small P_R(R) = C_A P_0 \rho_0 \frac{\cos\left(\alpha_{in}\right)}{R^2} ,
\end{equation}
which holds for objects with a diameter larger than the beam diameter at distance $R$ and requires additional information about (i) the incident angle $\alpha_{in}$
and (ii) the system constant $C_A$ independent of range and time.
However, the received laser power is typically corrected \cite{GeometricalIntensityModel}, as $C_A$ differs for each scanning layer due to different optics and beam divergences. Four different levels of intensity calibration can be reported according to~\cite{ReviewRadiometricProcessing}.
For the Velodyne HDL-S3D sensor used in our experiments, a beam divergence correction is applied following the sensor manual \cite{Velodyne}. This correction is defined as
\begin{equation}\label{eq:i}
    \small i = P_R + f_s \left\lvert f_o-\left(1-\frac{R}{R_{\text{max}}}\right)\right\rvert^2,
\end{equation}
where $f_s$ is the focal slope and $f_o$ is the focal offset.
The parameters for each laser are retrieved from the factory side calibration. Before applying the proposed simulation methods, we first retrieve the raw intensities by inverting this intensity calibration. 
In snowfall, the optical medium contains particles which are smaller than the beam diameter, so Mie scattering and the exact spatial distribution of the particles must be taken into account~\cite{Rasshofer_2011}.

\PAR{Pulse propagation in the presence of scattering particles} is described by a linear model introduced in~\cite{Rasshofer_2011}, which is valid for non-elastic scattering. This model expresses the range-dependent received power $P_R$ as a time-wise convolution between the time-dependent transmitted signal power $P_T$ and the impulse response $H$ of the optical system:
\begin{equation} \label{eq:LiDAR:equation}
    \small P_R(R) = C_A \int_0^{2R/c} P_T(t)\,H\left(R-\frac{ct}{2}\right) dt,
\end{equation}
with the time signature of the transmitted pulse given by
\begin{equation} \label{eq:pulse:sin2}
    \small
    P_T(t) = \left\{
    \begin{array}{rl}
        P_0\sin^2\left(\frac{\pi}{2\,\tau_H}t\right), & 0 \leq t \leq 2\,\tau_H, \\
        0 & \text{otherwise}.
    \end{array}
    \right.
\end{equation}
$\tau_H$ is the half-power pulse width, set to \unit[10]{ns} for the Velodyne HDL-S3D sensor. The impulse response $H$ can be factored into the impulse responses of the optical channel, $H_C$, and the target, $H_T$:
\begin{align}\label{eq:impulse:response:decomposition}
    \small H(R) =\ & H_C(R)\,H_T(R).
\end{align}
$H_C$ depends on the beam divergence, the overlap of transmitter and receiver described by $\xi(R)$ as well as the transmittance $T(R)$ of the medium through
\begin{align}\label{eq:impulse:response:channel}
    \small H_C(R) = \frac{T^2(R)}{R^2} \xi(R).
\end{align}
The transmittance $T(R)$ is equal to 1 in the part of the medium that is not occupied by snow particles, assuming absence of other scatterers. The overlap $\xi(R)$ can be geometrically derived from Fig.~\ref{fig:xi} as 
\begin{equation} \label{eq:xi}
    \small
    \xi(R) = \left\{
    \begin{array}{rl}
        0, & R \leq R_1 \\
        \frac{R-R_1}{R_2-R_1}, & R_1 < R < R_2 \\
        1, & R_2 \leq R.
    \end{array}
    \right.
\end{equation}

The impulse response of the target, $H_T$, allows us to model snow particles as we detail in the following.

\PAR{Scene reflection} defines the particle interaction with the laser pulse through $H_T$. For an extended solid target object we can write
\begin{equation}\label{eq:impulse:response:target:clear}
    \small H_T(R)=\rho_0\delta\left(R-R_0\right),
\end{equation}
with $\rho_0$ being the reflectivity of the object and $\delta$ the Dirac delta function. However, in snowfall, apart from the solid target object, the laser beam is also partially reflected by snow particles. 

We model snow particle $j$ as a spherical object with reflectivity $\rho_s$, diameter $D_j$ following the distribution introduced in~\cite{Gunn1958} and distance $R_j$ from the sensor, placed uniformly at random around the sensor so that it does not intersect with any other particle. The number of particles is chosen according to the snowfall rate, typically ranging in \unit[0-2.5]{mm/h} (see supplementary materials for more details). Particles can occlude each other and the target object, as illustrated in Fig.~\ref{fig:theory} (top). Thus, each particle $j$ reflects only a fraction $\theta_j/\Theta$ of the opening angle $\Theta$ of the beam, also letting a fraction $\theta_0/\Theta$ of the beam reach the target. Details on calculating the ratios $\theta_j/\Theta$ are given in the supplementary materials as well. 

\noindent
Assuming $D_j \ll c\tau_H$ for all $j$, we can write
\begin{equation} \label{eq:impulse:response:target:snow}
    \small H_T(R) = \frac{1}{\Theta} \left(\rho_0 \blank{1pt} \theta_0 \delta(R-R_0) + \rho_s\sum_{j=1}^n \theta_j\delta(R-R_j) \right), 
\end{equation}
with $\Theta=\theta_0 +\sum_{j=1}^n \theta_j$. 
Plugging \eqref{eq:pulse:sin2}, \eqref{eq:impulse:response:decomposition}, \eqref{eq:impulse:response:channel} and \eqref{eq:impulse:response:target:snow} into \eqref{eq:LiDAR:equation}, the received power in snowfall is 
\begin{equation} \label{eq:response:snow:decomposition}
    \small P_{R,\text{snow}}(R) = P_{R,\text{snow}}^0(R) + \sum_{j=1}^n P_{R,\text{snow}}^j(R),
\end{equation}
where
{\scriptsize
\begin{align}
    &P_{R,\text{snow}}^j(R) \nonumber \\
    &{=}\;\frac{C_A P_0 \rho_s \theta_j \xi(R_j)}{\Theta {R_j}^2} \displaystyle\int_0^{2\tau_H} \sin^2\left(\frac{\pi}{2\tau_H}t\right) \delta(R-\frac{ct}{2}-R_0) dt \nonumber \\
    &{=}\;\left\{\begin{array}{rl}
        \frac{C_A P_0 \rho_s \theta_j \xi(R_j)}{\Theta {R_j}^2} \sin^2\left(\frac{\pi(R-R_j)}{c\tau_H}\right), & R_j \leq R \leq R_j+c\tau_H \\
        0 & \text{otherwise.}
    \end{array}\right. \label{eq:response:snow:single:term}
\end{align}
}%
$P_{R,\text{snow}}^0(R)$ can be derived by substituting $(\theta_j,\,R_j,\,\rho_s)$ with $(\theta_0,\,R_0,\rho_0)$ on the right-hand side of \eqref{eq:response:snow:single:term}. 

The received power is thus a superposition of multiple echoes, each associated with an object (snow particle or target object), as depicted in Fig.~\ref{fig:theory} (bottom). Crucially, the magnitude of each echo depends on the angle $\theta_j$ and the inverse square of the distance $R_j$ of the respective object from the sensor. In this work, we retrieve the maximum peak of the received power as the LiDAR return.
Thus, if a peak owing to snow particles is higher than the peak associated to the target object, the true echo is missed and a cluttered point is added to the simulated point cloud at the range of the former peak. 

Otherwise the target object intensity is attenuated according to its occlusion percentage. Our complete snowfall simulation is presented in Algorithm~\ref{algo:LidarSnowfall}. In Fig.~\ref{fig:QualitativeResults} we show a winterly example scene, once augmented with our snowfall simulation and once with the one proposed in LISA~\cite{LISA}. 

\begin{figure}[!t]
    \centering\vspace{-0.5eM}
    \includegraphics[width=\linewidth]{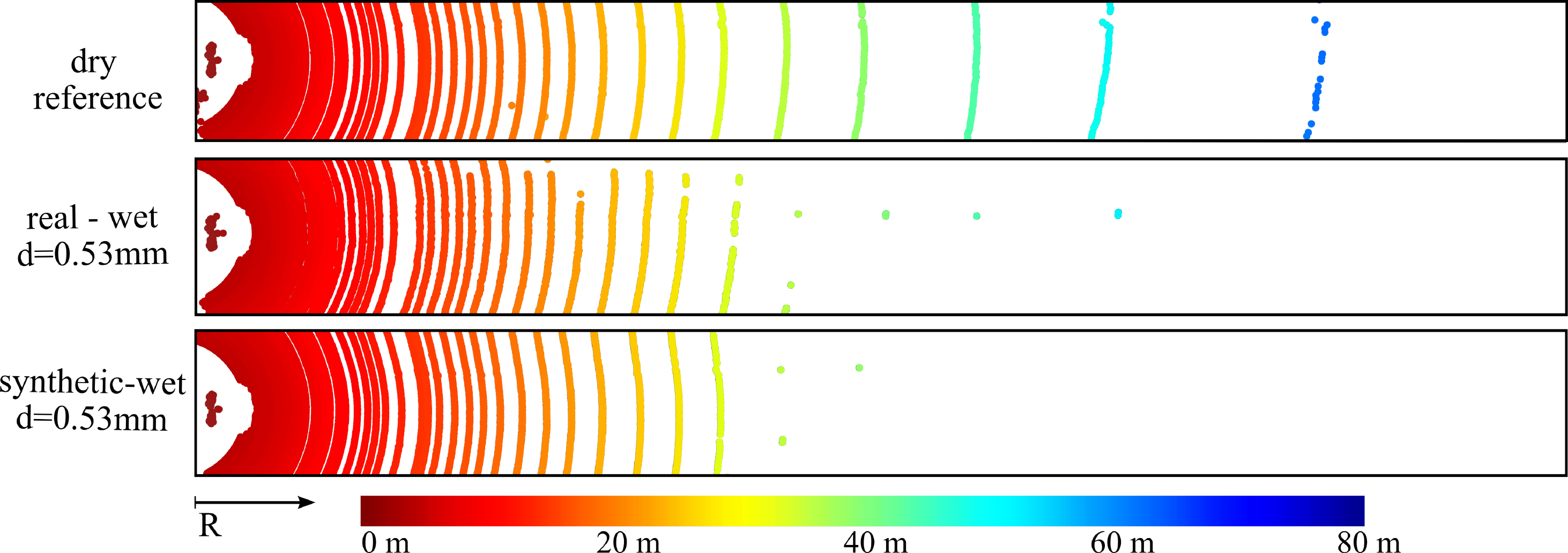}\vspace{-0.3eM}
    \caption{A real-world capture on a dry highway (top), a real-world capture with a water height of $d_w=\unit[0.53]{mm}$ (middle) and the synthesized road wetness from the clear reference (bottom).}
    \label{fig:RealAndSynWetRoadResults}\vspace{-0.7eM}
\end{figure}

\begin{algorithm}[!t]
  \caption{LiDAR snowfall simulation}\label{algo:LidarSnowfall}
  \begin{algorithmic}[1]
    \small \Procedure{snowfall}{$\mathbf{pc}, n_l, \tau_H, R_{\text{max}}, \Theta, r_s, \rho_s, \rho_0$}
        \small \For{$l$ in $n_l$}                                                                                                                                                               \scriptsize \Comment{for each layer $l$} 
            \small \State $\mathbf{pc}_{l} \gets \mathbf{pc}.$\scriptsize SELECT\small($\text{layer} = l$)
            \small \State $f_d, f_s, i_{\text{max}} \gets$ \scriptsize LOAD\_CALIB\small($l$)
            \small \State $f_o \gets (\frac{1 - f_d}{13100})^{2}$                                                                                                                               \scriptsize \Comment{focal offset~\cite{Velodyne}} 
            \small \State $\mathbf{s} \gets$ \scriptsize SAMPLE\_SNOWFLAKES\small($R_{\text{max}}, r_s$)                                                                                        \scriptsize \Comment{in 2D \cite{Gunn1958}} 
            \small \For{$\mathbf{p}$ in $\mathbf{pc}_{l}$}                                                                                                                                      \scriptsize \Comment{for each point in layer $l$} 
                \small \State $x, y, z, i \gets \mathbf{p}$
                \small \State $R_0 \gets {\| \mathbf{p} \|}_2$
                \small \State \pmb{t} $\gets$ \scriptsize GET\_PARTICLES\_IN\_BEAM\small($\mathbf{s}, x, y, R_0, \Theta$)                                                                         \scriptsize \Comment{in 2D}
                \small \If{len(\pmb{t}) $> 1$}                                                                                                                                                  \scriptsize \Comment{otherwise no interference} 
                    \small \State $\mathbf{P}_{R, \text{snow}} \gets \mathbf{0}$                                                                                                                       \scriptsize \Comment{initialize with zeros} 
                    \small \For{$i, R, \theta$ in \pmb{t}}                                                                                                                              \scriptsize \Comment{for each target}
                        \small \If{$R = R_0$}                                                                                                                                                   \scriptsize \Comment{original target}
                            \small \State $P_R \gets \scriptsize i - f_s \left\lvert f_o-\left(1-\frac{R}{R_{\text{max}}}\right)\right\rvert^2$  
                            \small \State $C_A P_0 \gets \frac{P_R}{\rho_0} {R_0}^2$                                                                                                              \scriptsize \Comment{follows from Eq.~\eqref{eq:response:snow:single:term}}
                        \small \Else                                                                                                                                                            \scriptsize \Comment{snowflake}
                            \small \State $P_R \gets \rho_s \blank{1pt} i_{\text{max}}$
                            \small \State $C_A P_0 \gets \frac{P_R}{\rho_0}$
                        \small \EndIf
                        \small \State \scriptsize $\mathbf{P}_{R, \text{snow}} \mathrel{+}=$ \scriptsize Eq.\eqref{eq:response:snow:single:term}\small($C_A P_0, R, \rho, \tau_H, \theta, \Theta$)
                    \small \EndFor
                    \small \State $P_R \gets \max(\mathbf{P}_{R, \text{snow}})$
                    \small \State $R^* \gets \argmax(\mathbf{P}_{R, \text{snow}}) - c \frac{\tau_H}{2}$
                    \small \State $i \gets \scriptsize P_R + i_{\text{max}} f_s \left\lvert f_o-\left(1-\frac{R^*}{R_{\text{max}}}\right)\right\rvert^2$   
                    \small \State $(x, y, z) \gets \frac{R^*}{R_0} \times (x, y, z)$
                    \small \State $\mathbf{p} \gets x, y, z, i$
                \small \EndIf
            \small \EndFor
        \small \EndFor
        \small \State \textbf{return} $\mathbf{pc}$
    \small \EndProcedure
  \end{algorithmic}
\end{algorithm}

\begin{figure}[!t]
    \centering
    \includegraphics[width=\linewidth]{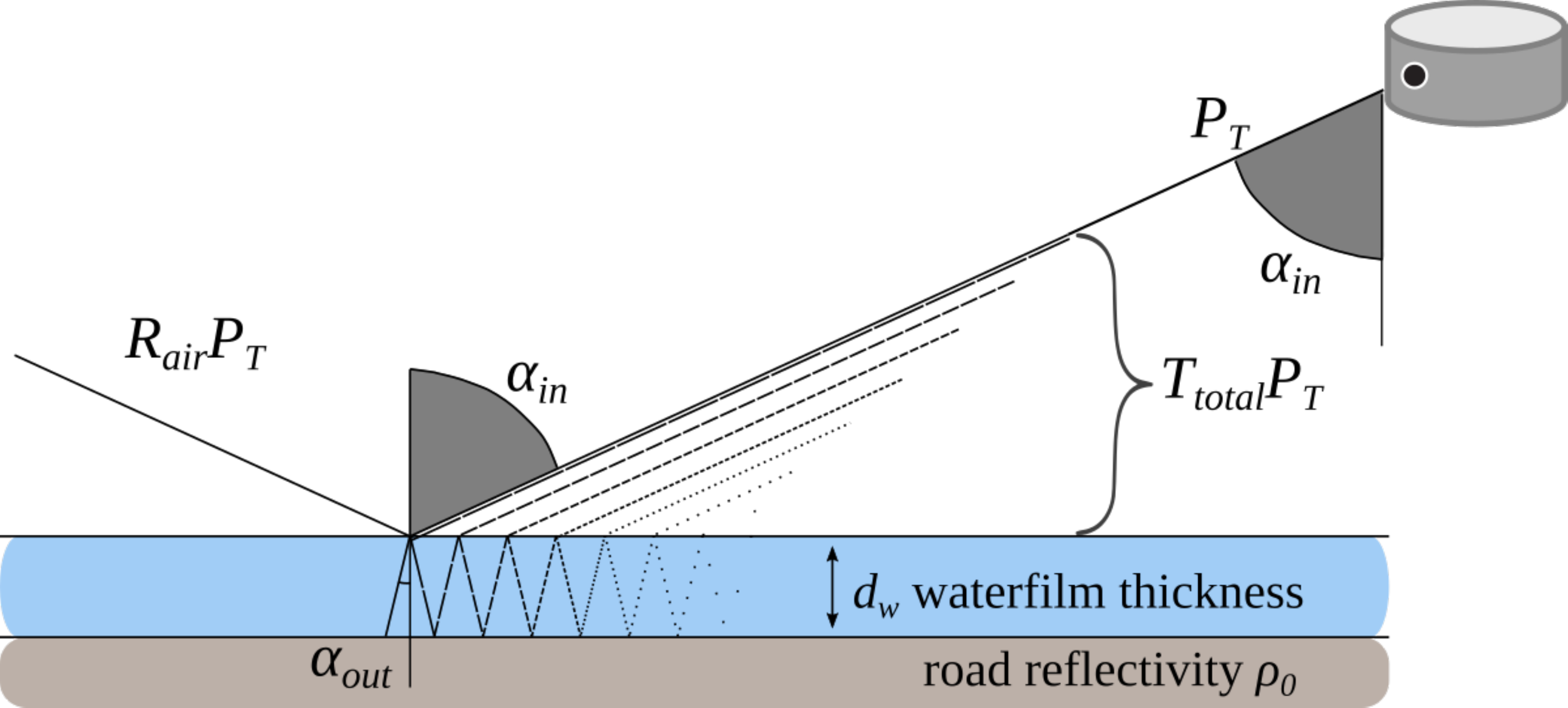}
    \caption{Visualization showing the geometrical optical model which describes the reflection on a wet road surface.}\label{fig:gemetricalWaterModel}\vspace{-1eM}
\end{figure}

\subsection{Wet Ground Model}
\label{sec:method:Groundwetness}

LiDAR readings are affected by the wetness of surrounding surfaces. Emitted light pulses are reflected specularly from wet ground leading to significantly attenuated laser echoes depending on the water height~\cite{WetAsphaltMeasurements}. Analysing the road wetness statistics of STF~\cite{STF} (given in the supplementary materials), it becomes apparent that wet roads occur together with adverse weather such as snowfall and are the main cause for lost points on road surfaces (see Fig.~\ref{fig:RealAndSynWetRoadResults}). 

To model the attenuation caused by road wetness, we apply geometric optics modeling single rays and their refraction on a thin layer of water illustrated in Fig.~\ref{fig:gemetricalWaterModel}. A qualitative example is shown in Fig.~\ref{fig:RealAndSynWetRoadResults} (bottom). We use the refractive indices $(n_\text{in}, n_\text{out})$ and angles $(\alpha_\text{in}, \alpha_\text{out})$ at the transition point. The angle $\alpha_\text{out}$ can be calculated based on Snell's law:
\begin{equation}
    \small \sin\left(\alpha_\text{out}\right)=\frac{n_\text{in}}{n_\text{out}}\sin\left(\alpha_\text{in}\right).
\end{equation}
The change in pulse amplitude is modeled by the Fresnel equations described by the perpendicular and parallel transmissions $t^\perp, t^=$ and reflections $r^\perp, r^=$ with respect to the ground, respectively (details in the supplementary materials).
Going from amplitude to transmitted power, we can deduce the power reflection $R^p_k$ and transmission $T^p_k$:
\begin{align}
    R^p_k = & \left(r^p_k\right)^2, \\
    T^p_k = & \frac{n_\text{in}\cos{(\alpha_\text{in}})}{n_\text{out}\cos{(\alpha_\text{out}})} \left(t_k^p\right)^2,
\end{align}
where $p \in (\perp, =)$ refers to the polarization.
We sum up all individual light rays which traverse back to the sensor, as shown in Fig.~\ref{fig:gemetricalWaterModel}, leading to the geometric series
\begin{equation}
    \small T_\text{total}^p = T_\text{air}^p\rho_0 T_\text{water}^p\sum_{k=0}^\infty (\rho_0 R_\text{water}^p)^k, 
\end{equation}
which can be simplified for $(\rho_0 R_{water}) < 1$ to
\begin{equation}\label{eq:RoadWetteningEquation}
    \small T_\text{total}^p=T_\text{air}^p\rho_0 T_\text{water}^p/(1-\rho_0 R_\text{water}^p).
\end{equation}
As the actual polarization is proprietary, we assume that the manufacturer optimized the polarization depending on the distance for best possible performance, implying
\begin{equation}
   \small T_\text{total} = \max\left(T_\text{total}^=, T_\text{total}^\perp\right).
\end{equation}

\begin{algorithm}[!t]
  \caption{LiDAR wet ground simulation}\label{algo:LidarGroundPlane}
  \begin{algorithmic}[1]
    \small \Procedure{wet\_ground}{$\mathbf{p}, i, \mathbf{w}, h, \epsilon_{g}, d_w, d_p, P_T, i_n$}
      \small \If{$|\mathbf{p}\cdot \mathbf{w} -h|< \epsilon_{g} $}
        \small \State $R_0 \gets \| \mathbf{p} \|_2$
        \small \State $\alpha_{in}  \gets  \arccos\left(\frac{\mathbf{p} \cdot \mathbf{w}}{R_0 \| \mathbf{w} \|_2}\right)$     
        \small \State $\rho_0 \gets \frac{i}{\cos\left(\alpha_\text{in}\right) P(R_0)}$ \Comment{\scriptsize{see Eq.~\eqref{eq:rho_0}}}\small
        \small \State $\gamma = \min\left(\max\left(d_w / d_p, 0\right), 1\right)$  \Comment{\scriptsize{see Eq.~\eqref{eq:gamma}}}\small
        \small \State $\rho_w \gets (1-\gamma)\cdot\rho_0 + \gamma\cdot T_\text{total}/\cos\left(\alpha_\text{in}\right)$                \Comment{\scriptsize{see Eq.~\eqref{eq:WeightingRho}}}\small
        \small \State $i \gets \rho_w \cos\left(\alpha_\text{in}\right) P_T(R_0)$  
        \small \If{$i>i_{n}\left(R_0\right)$}
            \small \State \textbf{return} $\textbf{p}$, $i$
        \small \Else
            \small \State \textbf{return} \textit{None}
        \small \EndIf
      \small \Else
        \small \State \textbf{return} $\textbf{p}$, $i$
      \small \EndIf
    \small \EndProcedure
  \end{algorithmic}
\end{algorithm}\vspace{-1eM}

Based on this formulation, we design Algorithm~\ref{algo:LidarGroundPlane}. A pre-processing step involves estimation of the ground plane normal $\mathbf{w}$ and intercept $h$ with RANSAC~\cite{Fischler1981RandomSC} as detailed in the supplementary materials.
These parameters allow to identify all points belonging to the ground
as well as the beam incident angle $\alpha_{in}$.
Then the modified intensity can be reconstructed from the measured ``dry'' intensity, taking into account that the beam divergence $\Theta$ has been corrected by a factory side calibration leading to $i \propto\rho_0 P_T(R) \cos\left(\alpha_{in}\right)$. Normalizing $i$ with the incident angle $i(R)/\cos{(\alpha_{in})}$ returns a linear correspondence between the measured intensities $i$ and the range $R$, such that we can approximate the power $P_T(R)$ and noise floor $i_{n}(R)$ linearly from the known given echoes.
Then we can obtain the reflectivity $\rho_0$ for each point $\mathbf{p}$:
\begin{equation}
    \small \rho_0 = i / \left( \cos\left(\alpha_{in}\right)P(R_0) \right). \label{eq:rho_0}
\end{equation}
These corrected reflectivities are augmented by weighing dry and wet reflections, assuming a road thread profile of depth $d_p$ filled with water of depth $d_w$, which yields
\begin{align}
   \small \gamma\left(d_w,d_p\right) =\ & \min\left(\max\left(d_w / d_p, 0\right), 1\right), \label{eq:gamma} \\
    \rho_w(\gamma) =\ & (1-\gamma)\cdot\rho + \gamma\cdot T_{total}/\cos\left(\alpha_{in}\right). \label{eq:WeightingRho}
\end{align}
Finally, the measured intensity $i$ is updated based on the modified reflectivity $\rho_w$ and the resulting point $\mathbf{p}$ is only kept if its intensity $i$ is greater than the noise floor $i_{n}$.
\section{Experiments}
\label{sec:results}

\begin{table*}
    \smaller
    \centering
    \setlength\tabcolsep{3pt}
    \begin{tabular*}{\linewidth}{l @{\extracolsep{\fill}} l rrrr | rrrr | rrrr }
    \textbf{Detection}              & \textbf{Simulation}   & \multicolumn{4}{c|}{\textbf{heavy snowfall} $\uparrow$}                                    & \multicolumn{4}{c|}{\textbf{light snowfall} $\uparrow$}                                    & \multicolumn{4}{c}{\textbf{clear weather} $\uparrow$}                               \\ 
    
    \textbf{method}                 & \textbf{method}       & \notsotiny \textbf{0-80m} & \notsotiny 0-30m  & \notsotiny 30-50m & \notsotiny 50-80m     & \notsotiny \textbf{0-80m} & \notsotiny 0-30m  & \notsotiny 30-50m & \notsotiny 50-80m     & \notsotiny \textbf{0-80m} & \notsotiny 0-30m  & \notsotiny 30-50m & \notsotiny 50-80m \\ 
    
    \hline \hline \noalign{\vskip 1mm}
    
                                    & \clear                & 39.69                     & 65.05             & 36.14             & 8.03                  & 41.13                     & \underline{69.24} & 39.72             & \textbf{11.68}        & 45.36                     & \textbf{72.34}    & 42.48             & 10.53             \\ 
                                    & \FOG                  & 38.19                     & 64.72             & 33.38             & 7.49                  & 39.82                     & 68.41             & 38.68             & 9.65                  & 43.37                     & 71.05             & 40.03             & 9.90              \\
                                    & \DROR                 & 38.57                     & 64.27             & 35.40             & 8.07                  & 39.33                     & 66.73             & 38.14             & 10.51                 & 41.44                     & 67.76             & 38.48             & 9.44              \\
    PV-RCNN~\cite{PV-RCNN}          & \LISA                 & 39.21                     & 64.21             & 35.34             & 8.64                  & \underline{41.60}         & 69.15             & \textbf{41.08}    & 11.15                 & 45.30                     & 71.06             & 42.86             & \underline{11.45} \\
    \noalign{\vskip 1mm} 
                                    & Ours-wet              & 40.03                     & 65.34             & 35.82             & \textbf{9.31}         & 41.07                     & 68.49             & 40.03             & 11.02                 & 44.81                     & 71.60             & 42.71             & 10.63             \\
                                    & Ours-snow             & \underline{41.61}         & \underline{67.44} & \textbf{37.47}    & 8.84                  & 41.20                     & 68.79             & 40.20             & 11.13                 & \underline{45.61}         & \underline{72.14} & \textbf{43.40}    & 11.21             \\
                                    & Ours-snow+wet         & \textbf{41.79}            & \textbf{68.39}    & \underline{37.14} & \underline{8.85}      & \textbf{41.79}            & \textbf{70.30}    & \underline{41.01} & \underline{11.28}     & \textbf{45.71}            & 71.88             & \underline{43.31} & \textbf{11.69}    \\
    
    \hline \noalign{\vskip 1mm} 
    
                                    & \clear                & 39.47                     & 65.14             & 36.29             & 6.83                  & 41.25                     & 69.12             & 39.86             & \textbf{11.81}        & \underline{45.19}         & \underline{72.33} & \textbf{43.20}    & 10.69             \\
                                    & \FOG                  & 40.06                     & 65.58             & 36.78             & 7.33                  & 41.10                     & 68.93             & 39.25             & 10.98                 & 44.46                     & 71.67             & 41.78             & \underline{10.84} \\
                                    & \DROR                 & 38.16                     & 64.97             & 33.23             & 6.83                  & 38.48                     & 66.93             & 35.68             & 9.97                  & 40.65                     & 67.94             & 36.85             & 8.45              \\
    VoxelRCNN-Car~\cite{VRCNN}      & \LISA                 & 39.06                     & 66.61             & 33.56             & 6.93                  & 40.68                     & 68.80             & 38.78             & 10.75                 & 45.03                     & 72.05             & 41.96             & 10.59             \\
    \noalign{\vskip 1mm}
    \notsotiny (single class method)& Ours-wet              & 40.28                     & \underline{67.37} & 36.03             & 6.95                  & 40.89                     & \underline{69.41} & 39.28             & 10.37                 & 43.98                     & 72.08             & 40.23             & 9.23              \\
                                    & Ours-snow             & \textbf{41.20}            & \textbf{68.27}    & \underline{37.18} & \textbf{7.90}         & \textbf{41.75}            & \textbf{70.22}    & \textbf{40.95}    & \underline{11.78}     & 44.52                     & \textbf{72.40}    & 42.23             & 10.39             \\
                                    & Ours-snow+wet         & \underline{40.76}         & 66.58             & \textbf{37.74}    & \underline{7.78}      & \underline{41.57}         & 68.94             & \underline{40.82} & 11.66                 & \textbf{45.20}            & 71.30             & \underline{42.84} & \textbf{11.23}    \\
    
    \hline \noalign{\vskip 1mm} 
    
                                    & \clear                & 38.68                     & 63.50             & 34.20             & 8.23                  & \underline{40.91}         & 68.36             & 39.96             & 10.98                 & 44.11                     & \underline{71.66} & 40.97             & 9.76              \\
                                    & \FOG                  & 38.82                     & 64.43             & 33.70             & \underline{8.69}      & 40.82                     & \underline{68.71} & \underline{40.05} & 10.48                 & 43.79                     & 70.34             & 41.22             & 10.77             \\
                                    & \DROR                 & 38.42                     & 64.47             & 34.31             & 8.27                  & 38.69                     & 65.62             & 37.59             & 10.26                 & 40.80                     & 67.62             & 36.61             & 8.69              \\
    CenterPoint~\cite{CP}           & \LISA                 & 38.11                     & 63.74             & 33.58             & 7.67                  & 40.26                     & 68.16             & 39.25             & \underline{11.20}     & \underline{44.70}         & 71.54             & 41.46             & \textbf{11.23}    \\
    \noalign{\vskip 1mm}
                                    & Ours-wet              & 39.03                     & 63.17             & \underline{36.08} & 7.69                  & 40.65                     & 68.59             & 39.58             & 10.87                 & 43.85                     & 70.20             & 41.66             & \underline{10.82} \\
                                    & Ours-snow             & \underline{39.81}         & \underline{64.75} & 35.77             & \textbf{9.06}         & 40.68                     & \underline{68.71} & 39.65             & 10.40                 & \textbf{44.76}            & \textbf{71.90}    & \textbf{42.91}    & 10.13             \\
                                    & Ours-snow+wet         & \textbf{40.14}            & \textbf{65.54}    & \textbf{37.20}    & 8.41                  & \textbf{41.23}            & \textbf{69.19}    & \textbf{40.07}    & \textbf{11.84}        & 44.33                     & 71.24             & \underline{41.49} & 10.02             \\
    
    \hline \noalign{\vskip 1mm} 
    
                                    & \clear                & 36.59                     & 63.50             & 30.17             & \textbf{6.86}         & 38.03                     & 65.83             & 35.95             & \textbf{9.39}         & \underline{42.81}         & \underline{70.27} & \underline{39.90} & 9.18              \\
                                    & \FOG                  & 35.98                     & 62.23             & \underline{30.94} & 6.39                  & 38.17                     & 66.07             & 37.12             & 8.63                  & 41.82                     & 67.44             & 39.73             & \underline{9.39}  \\
                                    & \DROR                 & 35.85                     & 65.36             & 27.99             & 6.13                  & 35.43                     & 63.50             & 32.87             & 7.95                  & 39.48                     & 66.92             & 35.18             & 8.61              \\
    Part-A²~\cite{PartA2}           & \LISA                 & 37.12                     & \underline{65.57} & 30.05             & 5.96                  & 38.04                     & 66.62             & 36.72             & 8.25                  & 41.92                     & 70.02             & 38.84             & 7.79              \\
    \noalign{\vskip 1mm}
                                    & Ours-wet              & 37.13                     & 65.02             & 30.46             & 6.14                  & \underline{38.29}         & 65.80             & \underline{38.12} & \underline{8.85}      & 41.92                     & 69.12             & 39.37             & 9.00              \\
                                    & Ours-snow             & \textbf{37.73}            & \textbf{66.44}    & 29.54             & 6.46                  & 38.23                     & \textbf{67.72}    & 36.02             & 7.41                  & \textbf{43.41}            & \textbf{71.00}    & \textbf{40.63}    & \textbf{9.74}     \\
                                    & Ours-snow+wet         & \underline{37.52}         & 64.97             & \textbf{31.65}    & \underline{6.48}      & \textbf{38.66}            & \underline{67.00} & \textbf{38.18}    & 8.43                  & 42.45                     & \underline{70.27} & 39.19             & 8.83              \\
    
    \hline \noalign{\vskip 1mm} 
    
                                    & \clear                & 36.68                     & 61.74             & 33.25             & 6.14                  & 39.04                     & \underline{66.90} & 39.72             & 9.28                  & \underline{41.79}         & \textbf{68.58}    & \textbf{40.34}    & 7.96              \\
                                    & \FOG                  & 36.56                     & 62.93             & 33.03             & 5.52                  & 38.37                     & 65.71             & 39.60             & 8.50                  & 41.28                     & 67.79             & 38.82             & 7.88              \\
                                    & \DROR                 & 36.14                     & 62.64             & 31.64             & 5.28                  & 36.31                     & 63.52             & 36.62             & 7.77                  & 39.08                     & 64.96             & 36.54             & 7.70              \\
    PointRCNN~\cite{PRCNN}          & \LISA                 & 36.68                     & 62.85             & 31.80             & 5.78                  & 38.08                     & 65.20             & 38.96             & 8.96                  & \textbf{41.80}            & \underline{68.17} & 39.79             & \textbf{8.35}     \\
    \noalign{\vskip 1mm}
                                    & Ours-wet              & 37.07                     & 62.92             & \textbf{33.76}    & \textbf{6.38}         & \textbf{39.46}            & 66.33             & \textbf{41.51}    & \textbf{9.69}         & 41.70                     & 67.41             & 39.59             & 8.05              \\
                                    & Ours-snow             & \textbf{37.59}            & \textbf{63.99}    & \underline{33.63} & \underline{6.15}      & 38.60                     & 65.18             & 39.20             & \underline{9.47}      & 41.43                     & 67.61             & \underline{39.89} & \underline{8.15}  \\
                                    & Ours-snow+wet         & \underline{37.51}         & \underline{63.77} & 33.01             & 5.90                  & \underline{39.15}         & \textbf{67.10}    & \underline{39.89} & 9.43                  & 41.34                     & 67.76             & 39.58             & 7.87              \\
    
    \hline \noalign{\vskip 1mm} 
    
                                    & \clear                & 36.08                     & 61.53             & 30.92             & 6.60                  & 37.77                     & \textbf{65.68}    & 36.06             & 9.80                  & 42.10                     & 67.82             & \underline{39.52} & 10.81             \\
                                    & \FOG                  & 36.08                     & 61.65             & 31.25             & 7.65                  & 37.31                     & 64.27             & 35.43             & \textbf{10.55}        & 42.34                     & \textbf{69.85}    & 39.20             & 10.13             \\
                                    & \DROR                 & 35.04                     & 60.72             & 28.79             & 7.88                  & 35.09                     & 62.24             & 32.09             & 8.85                  & 38.96                     & 64.74             & 35.50             & 9.76              \\
    SECOND~\cite{SECOND}            & \LISA                 & 35.90                     & 59.31             & \textbf{32.81}    & 7.44                  & \textbf{38.07}            & 64.38             & \textbf{38.47}    & \underline{10.32}     & 41.75                     & 67.01             & 38.24             & \textbf{11.47}    \\
    \noalign{\vskip 1mm}
                                    & Ours-wet              & 36.79                     & 61.18             & \underline{32.48} & \textbf{8.86}         & \underline{38.04}         & 64.73             & \underline{37.31} & 9.95                  & \underline{42.48}         & 69.32             & 39.28             & 10.58             \\
                                    & Ours-snow             & \underline{36.83}         & \underline{61.94} & 31.40             & \underline{8.50}      & 37.99                     & 65.40             & 36.70             & 9.59                  & \textbf{42.72}            & 69.16             & \textbf{40.13}    & \underline{10.86} \\
                                    & Ours-snow+wet         & \textbf{37.03}            & \textbf{64.00}    & 31.06             & 7.05                  & 37.11                     & \underline{65.42} & 35.75             & 8.69                  & 42.31                     & \underline{69.80} & 38.25             & 10.18             \\
    
    \hline \noalign{\vskip 1mm} 
    
                                    & \clear                & 30.85                     & 52.45             & 27.31             & 5.59                  & 34.09                     & 59.88             & 32.80             & 8.52                  & \underline{38.24}         & 64.02             & \underline{35.76} & \underline{8.01}  \\
                                    & \FOG                  & 30.39                     & 52.13             & 26.79             & 5.71                  & \underline{35.38}         & 60.81             & \underline{35.15} & \underline{9.60}      & 37.74                     & \underline{64.56} & 34.48             & 7.30              \\
                                    & \DROR                 & 29.32                     & \textbf{54.52}    & 21.88             & 4.82                  & 30.99                     & 57.17             & 28.43             & 6.95                  & 34.72                     & 60.59             & 30.34             & 6.72              \\
    PointPillars~\cite{PP}          & \LISA                 & 28.70                     & 49.78             & 24.98             & 5.63                  & 33.87                     & \underline{60.93} & 31.38             & 8.70                  & 37.92                     & 63.98             & 34.61             & 7.94              \\
    \noalign{\vskip 1mm}
                                    & Ours-wet              & \underline{31.58}         & 52.81             & \underline{28.64} & \underline{6.93}      & 34.58                     & 60.57             & 34.00             & 8.66                  & 38.10                     & \textbf{65.12}    & 34.31             & 7.75              \\
                                    & Ours-snow             & \textbf{32.94}            & \underline{54.21} & \textbf{29.79}    & \textbf{7.81}         & \textbf{35.96}            & \textbf{61.50}    & \textbf{35.67}    & \textbf{10.13}        & \textbf{39.25}            & 64.37             & \textbf{36.65}    & \textbf{8.80}     \\
                                    & Ours-snow+wet         & 31.38                     & 54.09             & 27.67             & 6.14                  & 34.18                     & 60.27             & 33.12             & 8.08                  & 38.17                     & 64.40             & 34.73               & 7.95            \\
    
    \hline \hline 
    
    \end{tabular*}\vspace{-0.5eM}
    \caption{Comparison of simulation methods for 3D object detection in snowfall on STF~\cite{STF}. We report 3D average precision (AP) of moderate cars on three STF splits: the heavy snowfall test split with 1404 samples, the light snowfall test split with 2512 samples and the clear-weather test split with 1816 samples. ``Ours-wet'': our wet ground simulation, ``Ours-snow'': our snowfall simulation, ``Ours-snow+wet'': cascaded application of our snowfall and wet ground simulation.}\vspace{-1eM}
    \label{table:snowfall}
\end{table*}

\begin{figure*}
    \centering
    \vspace{-0.5eM}
    \includegraphics[width=\textwidth]{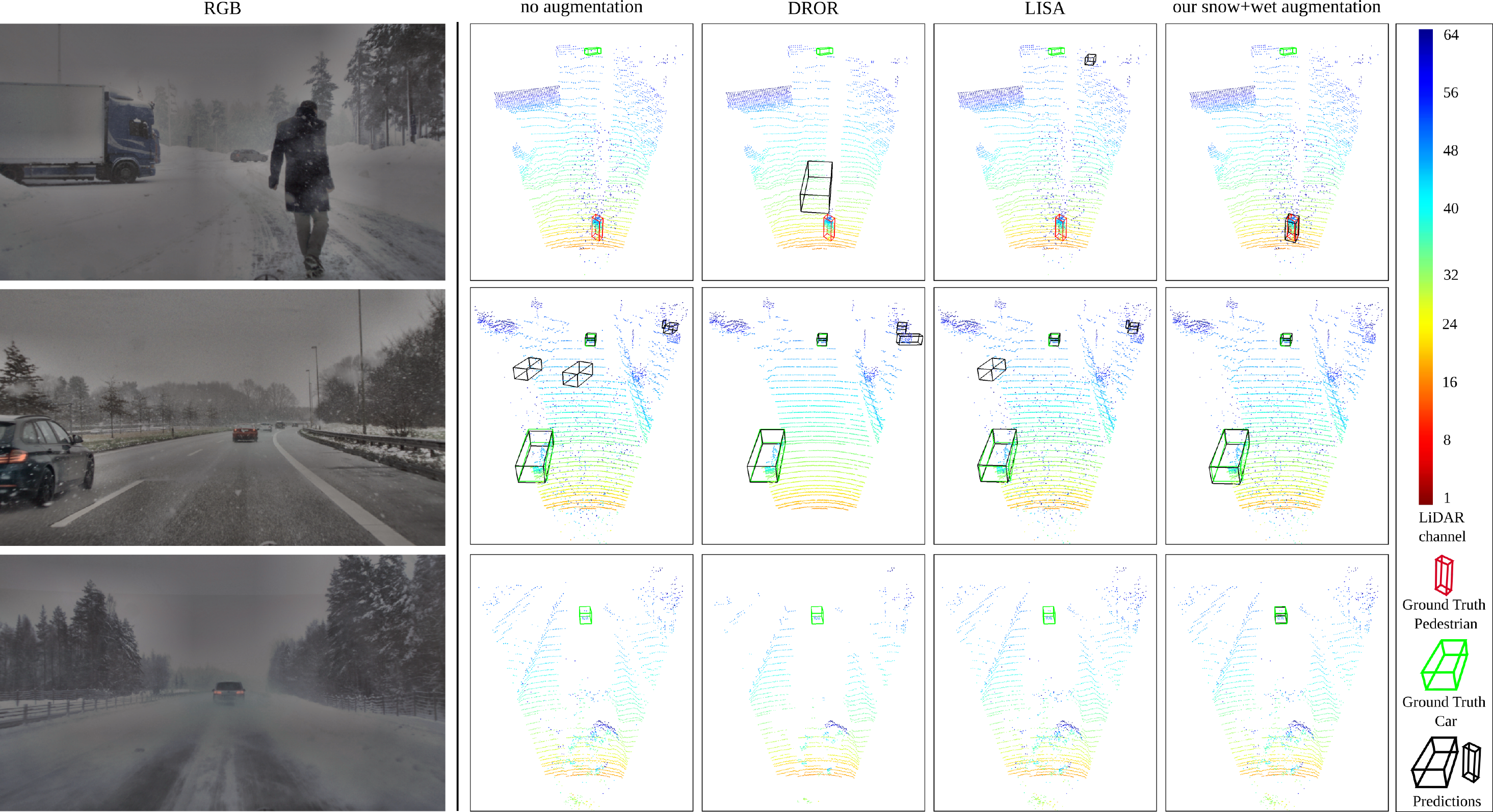}\vspace{-0.5eM}
    \caption{Qualitative comparison of PV-RCNN~\cite{PV-RCNN} on samples from STF~\cite{STF} containing heavy snowfall. The leftmost column shows the corresponding RGB images. The rest of the columns show the LiDAR point clouds with ground-truth boxes and predictions using the clear-weather baseline (``no augmentation''), DROR~\cite{DROR}, LISA~\cite{LISA}, and our fully-fledged simulation  (``our snow+wet augmentation'').}\vspace{-0.7eM} 
    \label{fig:results:qualitative}
\end{figure*}

\PAR{Dataset.} Our experiments are carried out on the STF dataset~\cite{STF}. It provides 12997 annotated samples with accurate 3D bounding boxes for object detection of cars, pedestrians, and cyclists in various weather conditions including light fog, dense fog and snow. Without denying the importance of the pedestrian and cyclist classes, in the main paper, we focus on the most dominant class, i.e.\ cars. In the supplementary materials we provide additional results. 
In total, 3469 frames in clear conditions can be used for training and 3916 frames in snowy conditions are provided.
We split the 3916 samples in the snow test set based on the intensity of snowfall into two different subsets, termed light snowfall and heavy snowfall, with 2512 and 1404 samples respectively. Inspired by~\cite{CADC}, we perform this split by leveraging the DROR algorithm~\cite{DROR}. The light snowfall split contains frames where DROR\cite{DROR} would filter 10-79 points from a 10$\times$2$\times$2m box in front of the ego vehicle, while the heavy snowfall contains at least 80 of such points within this box (further details in the supplementary materials). 

\PAR{Evaluation setting.}
For evaluation, we use the 3D object detection metrics defined in the KITTI evaluation framework~\cite{KITTI} and in \cite{yang2018pixor}. Specifically, \cite{yang2018pixor} introduces an extension to the KITTI metrics by reporting the results with respect to the object distance. Since the weather effects detailed in Sec.~\ref{sec:method} are distance-dependent, we opt for following their extension and report results in the intervals as in~\cite{Gated3D}. Additionally, we follow~\cite{Simonelli_2019_ICCV} and report average precision (AP) at 40 recall positions to provide a fair comparison. Other than that we use the typical overlap thresholds defined in~\cite{KITTI}.
To mitigate potential statistical fluctuations, we report for each experiment the average performance over three independent training runs.

\PAR{Baseline methods.}
In total, we investigate the effectiveness of our snowfall simulation scheme for seven well-known 3D object detection methods~\cite{PartA2, PV-RCNN, PRCNN, CP, PP, SECOND, VRCNN}. We compare our approach to a clear-weather baseline and two competing adverse weather simulation methods, one for fog~\cite{HahnerICCV21} and one for snowfall~\cite{LISA}. Additionally, we compare to denoising the point clouds using DROR~\cite{DROR}. To train the detection models, we use OpenPCDet~\cite{openpcdet2020} and follow the default training configurations for each method. All methods are trained from scratch.

\PAR{Data augmentation.}
We choose to apply our simulation(s) to every 10-th training sample, for which the snowfall rate is sampled from [0, 0.5, ..., 2.5]mm/h, and set the sensor constants $\tau_H=\unit[10]{ns}$, $R_{\text{max}}=\unit[120]{m}$, $\Theta=\unit[0.003]{rad}$, $\rho_s =0.9$ and $\rho_0 = \frac{1 \times 10^{-6}}{\pi}$. The exact same settings are used for~\cite{LISA}. For the wet ground simulation we use an exponential distribution and sample $d_w$ from the interval \unit[0.1-1.2]{mm}, while fixing $d_p$ to \unit[1.2]{mm} and setting $\epsilon_g$ to $0.5$m. 

\subsection{Quantitative Results}

We present the quantitative results in Table~\ref{table:snowfall}. In reading Table~\ref{table:snowfall}, the reader should first focus on the columns showing AP across the entire evaluation range of \unit[0-80]{m}. 
The main experimental finding from Table~\ref{table:snowfall} is that our 
full simulation including both the snowfall and the wet ground model (Ours-snow+wet) consistently improves the performance on the most challenging test case, i.e.\ heavy snowfall, for all methods by a significant margin compared to both the baseline approach as well as all competing simulation~\cite{HahnerICCV21, LISA} and denoising~\cite{DROR} methods. This improvement on heavy snowfall is particularly pronounced for the best-performing detection method, i.e.\ PV-RCNN~\cite{PV-RCNN}, for which our full simulation beats the clear-weather baseline by a notable 2.1\% in AP.

For PV-RCNN~\cite{PV-RCNN}, CenterPoint~\cite{CP} and Part-A²~\cite{PartA2}, our full simulation also delivers the best performance among all methods on light snowfall, showing that the benefit of our simulation extends to all snowfall intensities. Moreover, on the clear-weather test split, our snowfall simulation without wet ground modeling demonstrates the best performance among all competing approaches for six out of the seven detection methods, consistently improving upon the clear-weather baseline. This finding shows that using our snowfall simulation for training increases detection performance on severe snowy conditions, while not sacrificing but rather improving performance on clear weather as well. 

Using simulation methods designed for different adverse conditions, such as the fog simulation in~\cite{HahnerICCV21}, does not transfer well to snowfall as the respective physical models differ; performance of~\cite{HahnerICCV21} is slightly lower than the clear-weather baseline on both snowfall splits for most detection methods.
The application of DROR~\cite{DROR} as an enhancement step removing clutter points achieves among the lowest results, because it also removes several valid points, which do not belong to the snowfall clutter.

\subsection{Qualitative Results}

Qualitative results showing the proposed data augmentation scheme are presented in Fig.~\ref{fig:results:qualitative}. Here, PV-RCNN~\cite{PV-RCNN} is compared to the clear-weather baseline with no augmentation, DROR~\cite{DROR} and LISA~\cite{LISA}. In the first row, we see that the pedestrian inside the snowfall clutter can only be detected when our proposed data augmentation is applied during training. In the second row, additional false positives appear for all competing approaches. The bottom row shows a difficult highway scene with whirled-up snow dust. Our data augmentation approach generalizes well to this example, being the only method that detects the lead vehicle. Note also that in such a scenario with whirled-up snow dust, DROR~\cite{DROR} cannot remove the clutter completely.


\section{Conclusion}
\label{sec:conclusion}
In this work, we have introduced a novel method for realistic synthesis of winter scenes from clear LiDAR captures modeling snowfall and wet surfaces in a physically accurate way. 
Further, we have proven the effectiveness of the proposed algorithm, testing the augmentation with seven different 3D object detection methods and achieving consistent improvements of up to $2.1\%$ in AP in heavy snowfall. As future work, we envision the exploration of temporal cues for robust LiDAR-based 3D object detection.

\vspace{0.5cm}

\noindent
\textbf{Acknowledgements.} 
This work is funded by Toyota Motor Europe via the research project TRACE-Z\"urich. 
The work also received funding by the AI-SEE project with national funding from the Austrian Research Promotion Agency (FFG), Business Finland, Federal Ministry of Education and Research (BMBF) and National Research Council of Canada Industrial Research Assistance Program (NRC-IRAP). 
We also thank the Federal Ministry for Economic Affairs and Energy for support within “VVM-Verification and Validation Methods for Automated Vehicles Level  4 and 5”, a PEGASUS family project. 
Felix Heide was supported by an NSF CAREER Award (2047359), a Sony Young Faculty Award, and a Project X Innovation Award.
We thank Emmanouil Sakaridis for verifying our derivation of occlusion angles in our snowfall simulation.

{\small
\bibliographystyle{ieee_fullname}
\interlinepenalty=10000
\bibliography{refs}
}

\end{document}